\documentclass[11pt]{article}

\usepackage[final]{acl}

\usepackage{times}
\usepackage{latexsym}

\usepackage[T1]{fontenc}

\usepackage[utf8]{inputenc}

\usepackage{microtype}

\usepackage{inconsolata}

\usepackage{graphicx}

\usepackage{enumitem}
\usepackage{booktabs}
\usepackage{multirow}
\usepackage[table]{xcolor}

\definecolor{mygray}{gray}{0.9} 
\definecolor{myblue}{HTML}{ECF4FF}
\usepackage{amsmath}
\usepackage{tcolorbox}
\usepackage{stfloats}
\usepackage{algorithm}
\usepackage{algorithmic}
\usepackage{amssymb}

\newcommand{\normalrow}{\rowcolor{gray!30}}
\definecolor{deltaBg}{RGB}{220,230,255} 
\newcommand{\rowhighlight}{\rowcolor{deltaBg}}

\newcommand{\model}[1]{%
  \texttt{\hyphenchar\font=`\-
  #1}%
}

\definecolor{lxy}{RGB}{0,0,0}

\title{Teacher-Guided Policy Optimization for On-Policy Reasoning Distillation under Large Policy Divergence}

\author{
  \textbf{Xinyu Liu\textsuperscript{1}},
  \textbf{Kechen Jiao\textsuperscript{2}},
  \textbf{Chunyang Xiao},
  \textbf{Runsong Zhao\textsuperscript{1}},
  \textbf{Junhao Ruan\textsuperscript{1}},
  \\
  \textbf{Bei Li\textsuperscript{3}},
  \textbf{Jiahao Liu\textsuperscript{3}},
  \textbf{Qifan Wang\textsuperscript{4}},
  \textbf{Xin Chen\textsuperscript{3}},
  \textbf{Jingang Wang\textsuperscript{3}},
  \\
  \textbf{Chenglong Wang\textsuperscript{1}},
  \textbf{Tong Xiao\textsuperscript{1,5}\thanks{Corresponding authors.}},
  \textbf{Jingbo Zhu\textsuperscript{1,5}}
  \\
  \textsuperscript{1} \normalsize{School of Computer Science and Engineering, Northeastern University, China} \\
  \textsuperscript{2} \normalsize{Tsinghua University}
  \textsuperscript{3} \normalsize{Meituan}
  \textsuperscript{4} \normalsize{Meta AI} \\
  \textsuperscript{5} \normalsize{NiuTrans Research, Shenyang, China} \\
  \normalsize{\texttt{lxy1051493182@gmail.com}},
  \normalsize{\texttt{\{xiaotong, zhujingbo\}@mail.neu.edu.com}}
}

\begin{document}
\maketitle
\begin{abstract}

On-policy distillation (OPD) has emerged as a promising paradigm for reasoning-oriented post-training of large language models (LLMs), especially when combined with reinforcement learning from verifiable rewards (RLVR).
Existing OPD methods rely on reverse KL (RKL)-based teacher supervision over trajectories sampled from the student policy.
However, we identify a critical limitation: under large teacher--student policy divergence, RL-driven exploration often produces trajectories that the teacher assigns low probability, resulting in uninformative negative feedback.
To address this, we propose Teacher-Guided Policy Optimization (TGPO).
Unlike existing OPD methods that evaluate student-sampled actions using the teacher, TGPO directly guides next-token generation using the teacher's predicted distribution on student-generated contexts, yielding an on-policy forward-KL objective.
Combined with RLVR, this guidance steers exploration toward improved continuations.
Experiments on reasoning benchmarks show that TGPO outperforms existing RKL-based OPD methods and remains robust across different teacher models.
Furthermore, we find that teacher top-1 predictions suffice for effective guidance, without requiring the full teacher distribution.

\end{abstract}

\section{Introduction}

Reinforcement Learning with Verifiable Rewards (RLVR)~\cite{team2025kimi,guo2025deepseek} and knowledge distillation~\cite{qwen3technicalreport,xiao2026mimo} are two widely used approaches for improving the reasoning abilities of LLMs. 
RLVR enables scalable optimization from verifiable outcomes, but its reward signals are sparse and uniformly applied across all generated tokens, providing limited fine-grained feedback.
In contrast, knowledge distillation (KD) offers dense token-level supervision from a teacher model but relies on off-policy data.
Recently, \textbf{on-policy distillation (OPD)}~\cite{agarwal2024policy,lu2025onpolicydistillation,xu2025kdrl} has emerged as a promising paradigm that combines the advantages of both approaches.
Unlike conventional teacher-forced distillation, OPD trains the student on trajectories sampled from its own policy while leveraging teacher supervision signals.
By aligning training with the student-induced distribution, OPD alleviates the mismatch between training and inference and naturally complements RLVR-based reasoning optimization.

\begin{figure}[t!]
    \centering
    \includegraphics[width=0.48\textwidth]{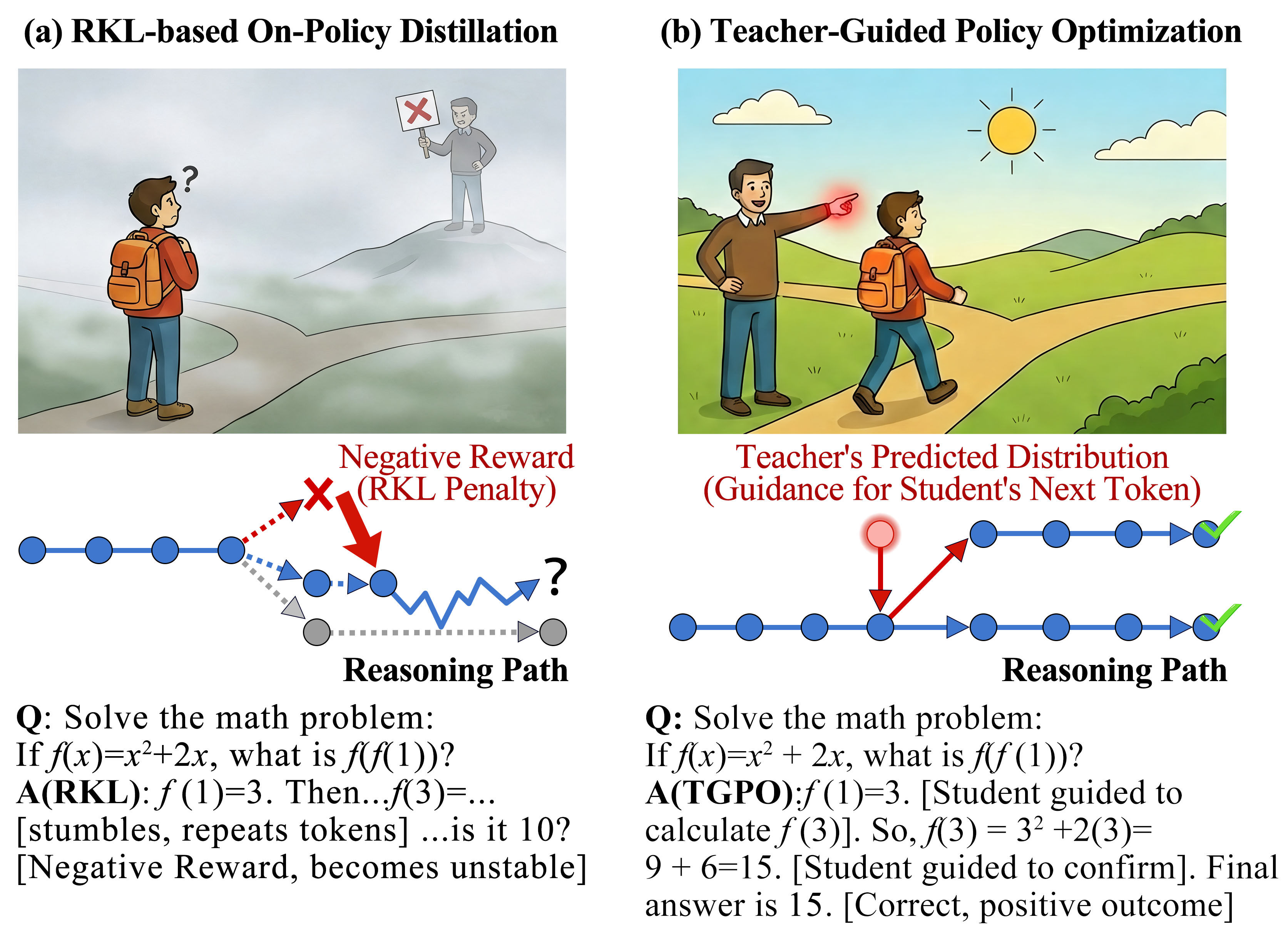}
    \caption{\textbf{RKL vs. TGPO. (a)} RKL relies on scalar rewards to penalize deviation. When the policy gap is significant, these penalties fail to provide directional information. \textbf{(b)} TGPO utilizes the teacher's predicted distribution as \textbf{guidance}, explicitly informing the student \textit{what} to generate next rather than \textit{what not} to generate.}
    \label{fig:intro_compare}
\end{figure}

Most existing OPD methods formulate teacher supervision through reverse KL (RKL)-based objectives, and prior work recommends RKL for compatibility with RL workflows~\cite{agarwal2024policy}. 
In practice, existing OPD methods often reduce teacher--student policy divergence before applying on-policy distillation. 
For example, prior work constructs teacher--student pairs within the same model family~\cite{agarwal2024policy}, employs self-teaching strategies~\cite{hubotter2026reinforcement,zhao2026self}, or introduces additional intermediate training stages to increase distribution overlap~\cite{lu2025onpolicydistillation,xu2025kdrl}.
These design choices suggest that RKL-based OPD methods implicitly rely on sufficient overlap between teacher and student trajectory distributions for effective supervision.
This raises a natural question: \emph{how does RKL-based OPD behave when the teacher and student distributions diverge substantially?}

As detailed in Section~\ref{sec:rkl-limitations}, we find that large policy divergence can severely degrade the effectiveness and stability of RKL-based supervision.
We attribute this failure primarily to the \emph{evaluative} nature of the RKL objective: it rewards trajectories preferred by the teacher while penalizing unlikely ones, but does not provide explicit guidance toward better continuations.
As a result, the student must rely on exploration to discover teacher-preferred trajectories.
When the student policy drifts far from the teacher distribution, the teacher assigns near-zero probability to many student-generated tokens, causing optimization to be dominated by uninformative negative feedback rather than useful directional guidance.
Such token-level penalties can further degrade the quality of sampled trajectories.
When combined with RLVR optimization (e.g., GRPO), this issue can destabilize optimization, as sampled groups become increasingly dominated by poor trajectories~\cite{le2025no}.


To address these limitations, we propose Teacher-Guided Policy Optimization (TGPO).
\textcolor{lxy}{As illustrated in Figure~\ref{fig:intro_compare}, unlike RKL-based objectives, which evaluate student-generated actions under the teacher policy, TGPO queries the teacher for its next-token distribution on student-generated contexts and optimizes the student's distribution toward it.
This yields an on-policy forward-KL supervision objective, which, for large vocabularies, can be approximated using the teacher's top-$k$ support or Monte-Carlo samples.
While GKD~\cite{agarwal2024policy} explores forward-KL-style OPD, the large-divergence regime we characterize, however, remains unexplored.
In this regime, combining our guidance objective with RLVR enables effective on-policy distillation.
We further show that teacher top-1 predictions suffice, matching the performance of larger top-$k$ or Monte-Carlo teacher targets. 
This eliminates the need for the full teacher distribution and reduces the guidance objective to a simple and efficient cross-entropy loss.
}
Based on this perspective, we make the following contributions:

\begin{itemize}[topsep=4pt, itemsep=4pt, parsep=2pt, leftmargin=*]

\item We analyze the limitations of RKL-based OPD and empirically show that its effectiveness depends on sufficient teacher--student distribution overlap.

\item We propose TGPO, which combines token-level teacher guidance with trajectory-level RLVR, enabling effective distillation under large teacher--student divergence.

\item Experiments on reasoning benchmarks show that TGPO improves the robustness of OPD under large teacher--student divergence, even outperforming the mixed-policy approach.

\end{itemize}

\section{RKL Limitations in LLM Distillation}

\label{sec:rkl-limitations}

In this section, we analyze the limitations of prior RKL-based on-policy distillation methods. We first formulate the RKL objective in Section~\ref{subsec:on_policy_distillation_framework}, then show why RKL-based supervision becomes unstable under large teacher--student distribution divergence in Section~\ref{subsec:low_reward}. Finally, Section~\ref{subsec:prelim_exp} provides empirical evidence supporting the analysis.

\subsection{RKL-Based On-Policy Distillation}
\label{subsec:on_policy_distillation_framework}

Given a prompt dataset $\mathcal{D}=\{x\}$, we aim to train a student policy $\pi_\theta(\cdot|x)$ to approximate a fixed, superior teacher policy $\pi_T(\cdot|x)$. On-policy distillation (OPD)~\cite{gu2023minillm,team2024gemma,agarwal2024policy} achieve this by minimizing the RKL divergence over student-generated responses $y$:

\begin{equation}
    \label{eq:rkl}
    \begin{split}
        \mathcal{J}_{\text{RKL}}(\theta) &= \mathbb{E}_{x \sim \mathcal{D}} D_{\text{KL}}(\pi_\theta || \pi_T) \\
        &= \mathbb{E}_{x \sim \mathcal{D}, y \sim \pi_\theta(\cdot|x)} \left[ \log \frac{\pi_\theta(y|x)}{\pi_T(y|x)} \right].
    \end{split}
\end{equation}

Unlike sequence-level or supervised KD, which rely on teacher-generated samples, the RKL objective takes expectations over responses sampled from the student policy itself. This on-policy formulation shares the same expectation structure as RL objectives, where optimization is also performed over trajectories sampled from the current policy. As a result, RKL-based distillation can be naturally interpreted within an RL framework.

Let $r(y)$ denote the reward assigned to a sampled sequence $y$. Standard RL objectives can then be written as:
\begin{equation}
    \label{eq:rl}
    \mathcal{J}_{\text{RL}}(\theta) = \mathbb{E}_{x \sim \mathcal{D}, y \sim \pi_\theta(\cdot|x)} \left[ r(y) \right].
\end{equation}

Comparing Eq.~\ref{eq:rkl} and Eq.~\ref{eq:rl}, minimizing $\mathcal{J}_{\text{RKL}}$ is equivalent to maximizing $\mathcal{J}_{\text{RL}}$ with intrinsic reward $r(y) = -\operatorname{sg}\!\left[\log \frac{\pi_\theta(y|x)}{\pi_T(y|x)}\right]$, enabling OPD to be naturally optimized within standard RL frameworks.

\begin{figure*}[t]
    \centering
    \includegraphics[width=0.98\textwidth]{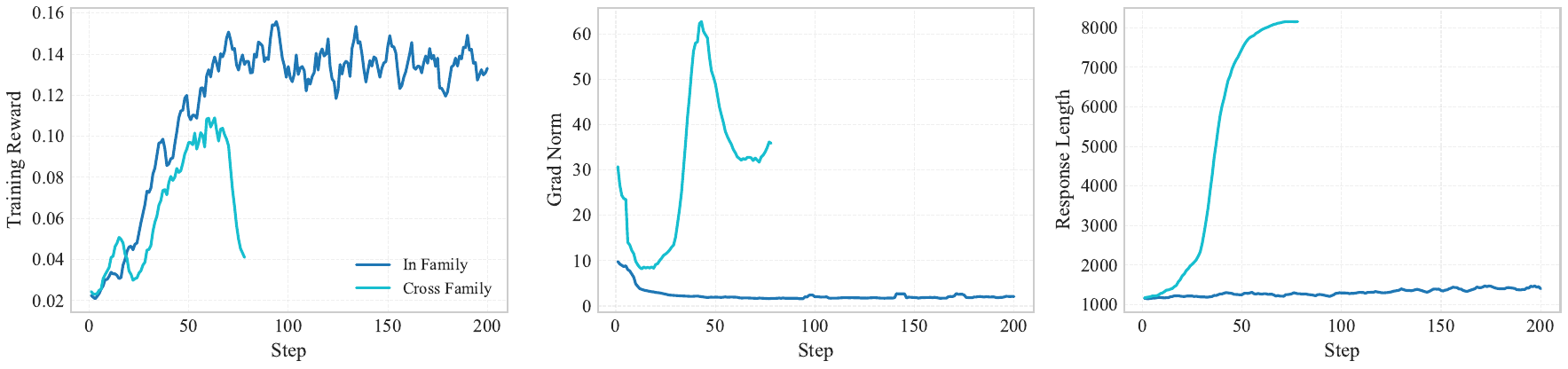} 
    \caption{Comparison of RKL distillation dynamics. We distill a \texttt{Qwen2.5-Math-1.5B} student using either an In-Family teacher (\texttt{Qwen2.5-Math-7B}) or a Cross-Family teacher (\texttt{Qwen3-30B-A3B}). While the In-Family setting converges stably, the Cross-Family setting exhibits catastrophic instability, characterized by sharp training score degradation \textbf{(Left)}, gradient norm spikes \textbf{(Middle)}, and unbounded response length growth \textbf{(Right)}.}
    \label{fig:sec2_compare}
\end{figure*}
\subsection{The Limitations of RKL-based Methods}
\label{subsec:low_reward}

Despite its simple formulation, the RKL objective introduces optimization challenges under large teacher--student distribution gaps. In this section, we analyze this issue by viewing RKL as an intrinsic reward within an RL framework, following recent studies on RKL-based OPD~\cite{xu2025kdrl,lu2025onpolicydistillation}.

Let $\rho(y)=\frac{\pi_\theta(y|x)}{\pi_T(y|x)}$ denote the density ratio. The intrinsic reward $-\log \rho(y)$ decreases monotonically with $\rho(y)$. Since trajectories $y$ are sampled from the student policy $\pi_\theta(\cdot|x)$, they are concentrated in regions where the student already assigns high probability. Consequently, the optimization dynamics mainly fall into two regimes:
\footnote{Because sampling is performed from $\pi_\theta$, trajectories with $\pi_\theta(y|x) \ll \pi_T(y|x)$ (i.e., $\rho(y)\ll1$) are rarely observed in practice. As a result, the student seldom receives strong positive rewards for trajectories favored by the teacher but not yet covered by the student policy.}:

\begin{itemize}[topsep=0pt, partopsep=0pt, itemsep=2pt, parsep=0pt, leftmargin=*]
    \item $\rho(y) \approx 1$ (\textbf{Consensus}): The generated trajectories lie within the teacher's high-probability support. In this case, the density ratio is close to one, leading to a near-neutral intrinsic reward ($-\log \rho(y)\approx0$).
    \item $\rho(y) \gg 1$ (\textbf{Rejection}): The student assigns high probability to trajectories that receive low probability under the teacher policy. This produces a large density ratio and a strong negative reward ($-\log \rho(y)\ll0$).
\end{itemize}

In the Consensus regime, successful rollouts are naturally reinforced by the RL algorithm. However, in the Rejection regime, the teacher functions merely as a punitive critic, providing only negative scalar feedback without guidance toward better actions. As a result, the student must explore the large action space through inefficient trial-and-error, which often leads to optimization stagnation. As illustrated in Figure~\ref{fig:intro_compare}(a), the lack of directional correction makes escaping the low-reward region computationally intractable.

Beyond the lack of directional guidance, RKL-based objectives also exhibit asymmetry in reward scaling. While the density ratio $\rho(y)$ becomes unbounded from above when $\pi_T(y|x) \to 0$, it is lower-bounded by the student's own probability $\pi_\theta(y|x)$. 
Because trajectories are sampled from the student policy, the ratio rarely falls far below 1.
As a result, negative penalties can dominate positive rewards by a large margin. 
This imbalance allows a single ``bad" sample to produce gradients that overwhelm the accumulated positive signals from ``good" samples, leading to unstable optimization. We provide a detailed analysis in Appendix~\ref{app:derivations}.

\subsection{Empirical Validation}
\label{subsec:prelim_exp}

We conjecture that training stability and performance degrade when the student frequently generates trajectories with high density ratios ($\rho(y) \gg 1$). To validate this, we train a \model{Qwen2.5-Math-1.5B} student under two configurations that induce different levels of teacher--student divergence\footnote{Detailed experimental settings and hyperparameters are provided in Appendix~\ref{app:exp_detail}.}:

\begin{itemize}[topsep=0pt, partopsep=0pt, itemsep=2pt, parsep=0pt, leftmargin=*]
    \item \textbf{In-Family Distillation:} \texttt{Qwen2.5-Math-7B} serves as the teacher. Since the teacher and student belong to the same model family and share similar training distributions, the resulting policy divergence is relatively small.
    \item \textbf{Cross-Family Distillation:} \texttt{Qwen3-30B-A3B} serves as the teacher.\footnote{We use the reasoning-oriented ``thinking'' MoE model as a proxy for a strong general-purpose model that differs substantially from the specialized math student.} Compared to the student, this teacher exhibits different reasoning behaviors and output distributions, leading to a larger policy divergence.\footnote{See Appendix~\ref{app:divergence} for quantitative measurements of teacher--student divergence across multiple metrics.}
\end{itemize}



\noindent \paragraph{Results.} Figure~\ref{fig:sec2_compare} shows markedly different training dynamics under the two settings. In the In-Family setting, RKL training converges steadily with consistent improvements in task reward, indicating that RKL provides effective supervision when the teacher and student are initially well aligned. In contrast, the Cross-Family setting exhibits severe instability. We observe three characteristic failure modes: (1) \textit{Performance Collapse}, where task rewards fail to improve consistently; (2) \textit{Exploding Gradients}, where persistently large gradient norms suggest that unbounded penalties destabilize optimization; and (3) \textit{Distributional Divergence}, where the student rapidly deviates from its initial distribution (e.g., pathological response length drift) after only $\sim$100 training steps. 
\textcolor{lxy}{We further verify that this instability is not specific to a particular RKL implementation: both the regularization form and a lower-variance top-100 distributional approximation exhibit the same failure, with detailed results in Appendix~\ref{app:rkl_variants}.}
When coupled with GRPO, such distributional divergence easily leads to groups dominated by low-reward samples.

\begin{figure*}[t!]
    \centering
    \includegraphics[width=0.78\textwidth]{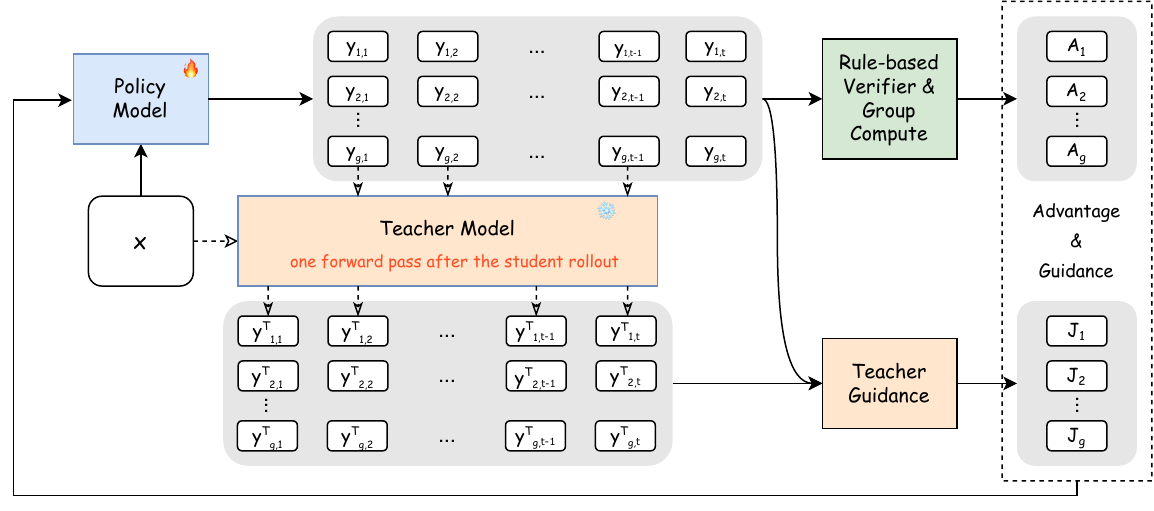} 
    \caption{Overview of the TGPO. The Policy Model generates a group of rollouts $\{y_i\}_{i=1}^g$ conditioned on input $x$. At each step, the Teacher Model provides dynamic token-level guidance by predicting the optimal target token $y^T$ based on the student's current prefix. This dense guidance signal ($J$) complements the outcome-based advantage ($A$) derived from the Rule-based Verifier to update the policy.}
    \label{fig:tgpo_frame}
\end{figure*}

\section{Teacher-Guided Policy Optimization}
\label{sec:method}
To address the limitations discussed in Section~\ref{sec:rkl-limitations}, we propose \textbf{Teacher-Guided Policy Optimization (TGPO)}, an on-policy distillation algorithm for reasoning-oriented LLM training. Instead of using the teacher for evaluative supervision, TGPO reformulates teacher feedback as directional guidance for policy optimization.

\subsection{Guidance on Student Trajectories}
\label{subsec:guidance}
\textcolor{lxy}{Similar to RKL-based methods, TGPO remains fully on-policy and operates on trajectories sampled from the student policy $\pi_\theta$. Given an input $x$, the student generates a trajectory
$y \sim \pi_\theta(\cdot \mid x)$, where each token $y_t$ is sampled conditioned on the input $x$ and the generated prefix $y_{<t}$. Let $s_t=(x,y_{<t})$ denote the student-visited state at step $t$. At each state, TGPO provides directional guidance by optimizing the forward KL,}
\textcolor{lxy}{
\begin{equation*}
\label{eq:forward_kl}
\begin{aligned}
& D_{\mathrm{KL}}(\pi_T(\cdot\mid s_t)\,\|\,\pi_\theta(\cdot\mid s_t)) \\
&= - H(\pi_T(\cdot\mid s_t)) - \sum_{v} \pi_T(v\mid s_t) \log \pi_\theta(v\mid s_t). 
\end{aligned}
\end{equation*}}

\textcolor{lxy}{Since the teacher is fixed during optimization, its entropy is constant, making the forward KL optimization-equivalent to its cross-entropy term. However, computing the full teacher distribution over a vocabulary of $\sim$151K tokens is expensive. We therefore use the teacher's top-$k$ support as the target.\footnote{See Appendix~\ref{app:topk_coverage} for top-$k$ coverage analysis.} Let $\mathcal{V}_k(s_t)$ denote its $k$ most probable next tokens, and let $\bar{\pi}_T(\cdot\mid s_t)$ denote the distribution renormalized over this support. The resulting state-level guidance loss is
\begin{equation}
\label{eq:guidance_token}
\ell_{\mathrm G}^{(k)}(s_t)
=
-\sum_{v\in\mathcal V_k(s_t)}
\bar\pi_T(v\mid s_t)
\log\pi_\theta(v\mid s_t).
\end{equation}
Aggregating this loss over student trajectories gives
\begin{equation}
\label{eq:guidance_general}
\mathcal J_{\mathrm G}^{(k)}
=
\mathbb E_{x\sim\mathcal D,\;y\sim\pi_\theta(\cdot\mid x)}
\left[
\sum_{t=1}^{|y|}
\ell_{\mathrm G}^{(k)}(s_t)
\right].
\end{equation}
Setting $k=1$ reduces the teacher target to its most likely next token,
\begin{equation}
\label{eq:guidance}
\mathcal J_{\mathrm G}
=
\mathbb E_{x \sim \mathcal{D},\, y \sim \pi_\theta(\cdot\mid x)}
\left[
-\sum_{t=1}^{|y|}
\log\pi_\theta(y_t^T\mid s_t)
\right],
\end{equation}
where $y_t^T=\arg\max_{v\in\mathcal V}\pi_T(v\mid s_t)$.}

\textcolor{lxy}{Our experiments show that larger top-$k$ or Monte-Carlo-sampled teacher targets yield no consistent gain over the top-1 target (Appendix~\ref{app:fkl_ablation}). We thus use Eq.~\ref{eq:guidance} as TGPO's guidance objective, which requires only the teacher's argmax token.}

Unlike RKL-based objectives, which evaluate the teacher's likelihood of actions sampled by the student, our guidance objective directly provides a teacher-preferred continuation at each student-visited state. This gives the student explicit directional supervision toward promising regions of the trajectory space that may be difficult to discover through exploration alone. Mechanistically, our objective resembles the teacher-forcing loss used in SFT, but differs in the trajectory distribution: the prefixes $s_t$ are generated by the student rather than taken from static ground-truth trajectories. 
\textcolor{lxy}{Thus, the teacher dynamically corrects the student on its own visited states, providing a DAgger-style on-policy correction signal~\cite{ross2011reductionimitationlearningstructured}.}

\subsection{Integrating Guidance into GRPO}
\label{sec:integration}

Since Eq.~\ref{eq:guidance} is fully on-policy, it can be naturally integrated into RLVR methods such as Group Relative Policy Optimization (GRPO)~\cite{shao2024deepseekmath}.
Given $x \sim \mathcal{D}$, the policy $\pi_\theta$ samples a group of $G$ trajectories $\{y_i\}_{i=1}^G$.
Following recent work~\cite{yu2025dapo,he2025skywork,liu2503understanding}, we omit the explicit KL regularization term with respect to a reference policy. The GRPO objective is defined as:
\begin{equation*}
    \mathcal{J}_{\text{RL}} = \mathbb{E}_{x \sim \mathcal{D}, \{y_i\}_{i=1}^{G} \sim \pi_{\theta}} \left[ \frac{1}{Z} \sum_{i=1}^{G} \sum_{t=1}^{|y_i|} \rho_{i,t}(\theta) A_i \right]
\end{equation*}
where $Z = \sum_{i}|y_i|$ normalizes by the total number of generated tokens, $\rho_{i,t}(\theta) = \frac{\pi_{\theta}(y_{i,t} \mid x, y_{i,<t})}{\pi_{\theta_{\text{old}}}(y_{i,t} \mid x, y_{i,<t})}$ denotes the importance sampling ratio, and $A_i = \frac{r_i - \mu}{\sigma}$ denotes the normalized group advantage.

Existing OPD methods typically combine distillation and RLVR signals through either reward shaping or differentiable teacher regularization.
The main difference is whether teacher feedback is treated as a scalar reward on sampled trajectories or as a direct optimization target toward teacher-preferred continuations.

For TGPO, we adopt the latter formulation. 
Reward shaping is less suitable in our setting because GRPO updates the sampled student token $y_{i,t}$, while the guidance signal is defined on the teacher target token $y_{i,t}^{T}$. 
Using the guidance score as a scalar reward therefore introduces a mismatch between the optimized action and the supervised target. 
Instead, we directly optimize the likelihood of teacher targets conditioned on student-generated trajectories:
\textcolor{lxy}{\begin{equation}
\label{eq:differentiable_regularization}
\mathcal{J}_{\text{TGPO}}
=
\mathcal{J}_{\text{RL}}
-
w \mathcal{J}_{\text{G}},
\end{equation}}
where $w$ controls the strength of teacher guidance.

Strong guidance is useful in the early stage of training, but overly rigid supervision may later restrict exploration. 
To balance imitation and exploration, we linearly decay the guidance weight during training:
$w_t = \max(w_{\text{init}} - \delta \cdot t, 0)$, where $w_{\text{init}}$ is the initial guidance weight, $t$ is the current training step, and $\delta$ is the decay rate. 
This schedule gradually shifts training from teacher-guided optimization toward pure reward-driven optimization. We use the annealed formulation as the default TGPO setting, and refer to the variant with a fixed guidance weight as \textbf{TGPO w/o annealing}.

\begin{table*}[t]
\centering
\resizebox{\textwidth}{!}{%
\begin{tabular}{lccccc>{\columncolor{yellow!20}}c|ccc>{\columncolor{cyan!20}}c}
\toprule
\multirow{2}{*}{\textbf{Model}} & \multicolumn{6}{c}{\textbf{In-Distribution Performance}} & \multicolumn{4}{c}{\textbf{Out-of-Distribution Performance}} \\
\cmidrule(lr){2-7} \cmidrule(lr){8-11}
 & \textbf{AIME 24/25} & \textbf{AMC} & \textbf{MATH-500} & \textbf{Minerva} & \textbf{Olympiad} & \textbf{Avg.} & \textbf{ARC-c} & \textbf{GPQA}$^{*}$ & \textbf{MMLU-Pro} & \textbf{Avg.} \\
\midrule
\normalrow
\multicolumn{11}{c}{Original Models} \\
Qwen2.5-Math-7B 
  & 11.5/4.9 & 31.3 & 43.6 & 7.4 & 15.6 & 19.0 & 18.2 & 11.1 & 16.9 & 15.4 \\
Qwen3
  & 59.5/49.8 & 85.3 & 96.0 & 52.9 & 68.0 &  68.6    & 94.1 & 65.2 & 80.0 & 79.8    \\
Qwen3-8192
  & 25.1/17.4 & 52.2 & 86.2 & 47.4 & 47.7 & 46.0     & 93.8 & 49.0 & 76.5 & 73.1   \\
\midrule
\normalrow
\multicolumn{11}{c}{Off-Policy and Mixed-Policy Methods} \\
SFT
    & 12.9/15.1 & 45.3 & 80.4 & \textbf{42.3} & 41.0 & 39.5 & 73.1 & 20.2 & 44.9 & 46.1 \\
LUFFY
    & 19.6/14.9 & 57.6 & \underline{83.6} & 38.6 & \textbf{51.9} & \underline{44.4} & 80.1 & \textbf{38.9} & \textbf{50.1} & \underline{56.4} \\
\midrule
\rowhighlight
\multicolumn{11}{c}{On-Policy Methods} \\
SimpleRL-Zero    
    & 27.0/6.8 & 54.9 & 76.0 & 25.0 & 34.7 & 37.4 & 30.2 & 23.2 & 34.5 & 29.3 \\
PRIME-Zero
    & 17.0/12.8 & 54.0 & 81.4 & 39.0 & 40.3 & 40.7 & 73.3 & 18.2 & 32.7 & 41.4 \\
Oat-Zero
    & \textbf{33.4}/11.9 & 61.2 & 78.0 & 34.6 & 43.4 & 43.7 & 70.1 & 23.7 & 41.7 & 45.2 \\
GRPO++
    & 19.5/15.8 & 58.3 & 82.2 & 37.5 & 47.3 & 43.4 & 77.4 & 32.3 & 46.9 & 52.1 \\
KDRL
    & 17.2/14.4 & 55.8 & \underline{83.6} & 36.0 & 43.4 & 41.7 & 78.4 & 35.4 & 46.9 & 53.6 \\
OP Distill
    & 5.7/4.5 & 29.9 & 64.0 & 23.2 & 27.1 & 25.7 & 26.1 & 6.1 & 23.0 & 18.4 \\
\midrule
\textbf{TGPO w/o annealing}
    & 20.1/\underline{16.0} & \underline{58.6} & \underline{83.6} & 37.9 & 48.1 & 44.1 & \underline{81.2} & \underline{37.9} & \underline{48.9} & 56.0 \\
\textbf{TGPO}
    & \underline{21.1}/\textbf{17.9} & \textbf{60.2} & \textbf{84.4} & \underline{40.4} & \underline{49.8} & \textbf{45.6} & \textbf{82.8} & 37.4 & \textbf{50.1} & \textbf{56.8} \\

\midrule
\normalrow

\end{tabular}}
\caption{
In-distribution and out-of-distribution performance based on Qwen2.5-Math-7B. We primarily benchmark against on-policy reasoning baselines, while also including off-policy and mixed-policy methods for comparison. The teacher model employed is Qwen3-30B-A3B (Qwen3); we additionally report its performance with a maximum generation length of 8192 tokens (Qwen3-8192). All models are evaluated under a unified setting. Bold indicates the best result, and \underline{underline} indicates the second best (excluding the teacher model).}
\label{tab:main_results}
\end{table*}

\section{Experimental Setup}
\label{sec:experiments}

\paragraph{Model and Dataset Construction.} 
Following previous work~\cite{yan2025learning,liu2503understanding,zeng2025simplerl}, we adopt \texttt{Qwen2.5-Math-7B}~\cite{yang2024qwen25mathtechnicalreportmathematical} as our default base model. 
We adopt \texttt{Qwen3-30B-A3B}~\cite{qwen3technicalreport} as the teacher model, aligning with the Cross-Family setting described in Section~\ref{subsec:prelim_exp}. 
We use \texttt{OpenR1-Math-46k-8192}~\cite{yan2025learning}, a subset of \texttt{OpenR1-Math-220k}~\cite{openr1}, as the training prompt set. 
To enable direct comparison with off-policy and mixed-policy methods, we sample teacher responses for \texttt{OpenR1-Math-46k-8192} and filter incorrect outputs using Math-Verify\footnote{\url{https://github.com/huggingface/Math-Verify}}.
This process yields 35k prompts with corresponding off-policy reasoning traces.\footnote{These traces are used only for off-policy and mixed-policy methods, while TGPO requires prompts only.}
We further evaluate TGPO with \texttt{Qwen2.5-Math-1.5B} as the student model to study performance under a larger teacher--student gap. Additional details are provided in Appendix~\ref{app:1d5b_exp}.

\paragraph{Benchmarks and Metrics.}
We assess performance across six widely-adopted mathematical reasoning benchmarks: AIME24, AIME25, AMC~\cite{li2024numinamath}, Minerva~\cite{lewkowycz2022solving}, OlympiadBench~\cite{he2024olympiadbench}, and MATH500~\cite{hendrycks2021measuring}. 
For AIME24, AIME25, and AMC, we report \textbf{avg@32} due to their relatively small evaluation sets; for the remaining benchmarks, we use standard \textbf{pass@1}.
To evaluate out-of-distribution generalization beyond mathematics, we additionally report results on ARC-c~\cite{clark2018think}, GPQA-Diamond~\cite{rein2024gpqa} (denoted as GPQA$^{*}$), and MMLU-Pro~\cite{wang2024mmlu}.
During inference, we use a sampling temperature of $0.6$. 
We also shuffle multiple-choice options to reduce position bias and mitigate potential data contamination.

\paragraph{Baseline Methods.}
We compare TGPO against both on-policy reasoning baselines and off-policy/mixed-policy methods.
The on-policy baselines fall into two categories: RKL-based methods and pure RLVR methods.
For RKL-based approaches, we include OP Distill~\cite{lu2025onpolicydistillation}, which uses the RKL log-ratio as the advantage signal, and KDRL~\cite{xu2025kdrl}, which augments the GRPO objective with RKL regularization.
We further compare against four pure RLVR variants: 
(1) SimpleRL-Zero, trained with standard rule-based rewards; 
(2) Oat-Zero~\cite{liu2503understanding}, which adopts Dr.GRPO for simplified advantage computation and loss normalization; 
(3) PRIME-Zero~\cite{cui2025process}, which derives implicit process rewards from policy rollouts and outcome labels; 
and (4) GRPO++, which removes the explicit KL penalty and introduces token-level supervision.
For completeness, we also report results for two off-policy or mixed-policy methods:
(1) SFT, fine-tuned on teacher-sampled responses, 
and (2) LUFFY~\cite{yan2025learning}, which incorporates teacher-sampled trajectories as auxiliary supervision during RLVR training.
Detailed training configurations are provided in Appendix~\ref{app:exp_detail}.

\section{Experimental Results}
\label{sec:results}

\subsection{Main Results}

Table~\ref{tab:main_results} reports results on both in-distribution (ID) math tasks and out-of-distribution (OOD) reasoning benchmarks. 
TGPO w/o annealing achieves better performance than on-policy methods, while remaining competitive with the strong mixed-policy baseline LUFFY.
TGPO further improves over the no-annealing variant and achieves the best average performance on both ID and OOD benchmarks, highlighting the benefit of gradually annealing teacher guidance strength.

On ID benchmarks, TGPO improves over KDRL by 3.9 points (45.6 vs.\ 41.7). 
It also avoids the training collapse observed in OP Distill, indicating more stable optimization under on-policy exploration. 
Beyond on-policy distillation methods, TGPO also outperforms strong baselines from other training paradigms, including LUFFY (44.4) and the RLVR baseline GRPO++ (43.4). 
On OOD benchmarks, TGPO achieves the highest average score of 56.8, improving over SFT by 10.7 points (56.8 vs.\ 46.1). 
It also exceeds LUFFY on challenging reasoning tasks such as ARC-c (82.8 vs.\ 80.1), suggesting that teacher-guided on-policy training improves generalization to unseen reasoning tasks.

\begin{figure*}[t!]
    \centering
    \includegraphics[width=\linewidth]{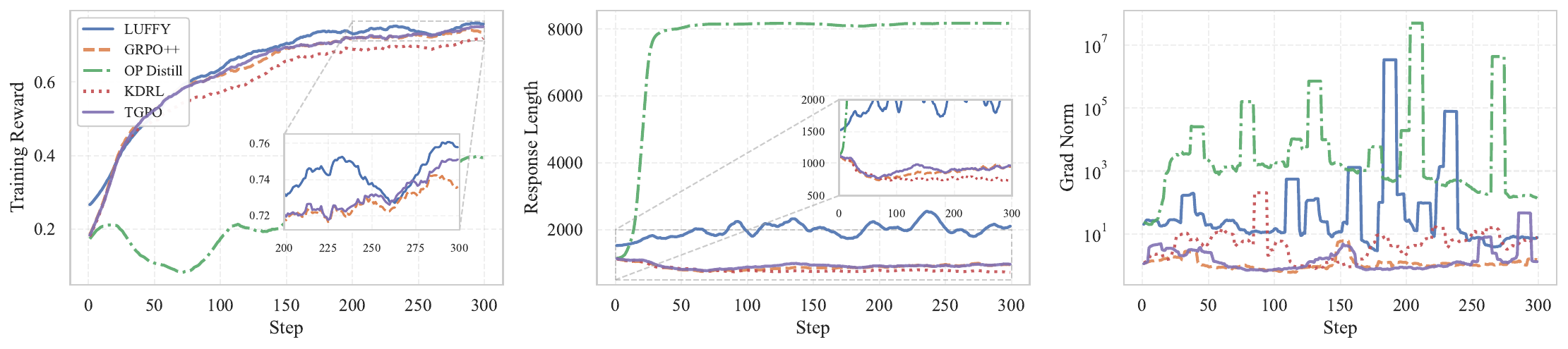}
    \caption{Training Dynamics Analysis. 
    \textbf{(Left)} Training reward. TGPO demonstrates robust growth and convergence compared to RKL-based methods (i.e., KDRL, OP Distill). 
    \textbf{(Middle)} Response length. TGPO avoids OP Distill's length explosion and aligns with GRPO++'s stability. 
    \textbf{(Right)} Gradient norm. TGPO shows stable optimization compared to the high variance in OP Distill, KDRL and LUFFY.}
    \label{fig:train_dynamics}
\end{figure*}

\subsection{Training Dynamics and Stability Analysis}

We analyze the training dynamics of TGPO and several baselines (GRPO++, KDRL, OP Distill, and LUFFY) using three metrics: training reward, response length, and gradient norm. 

Figure~\ref{fig:train_dynamics} shows that TGPO maintains stable training dynamics compared to RKL-based methods (OP Distill and KDRL).
Specifically, OP Distill exhibits early reward collapse (Left), severe response length explosion (Middle), and large gradient fluctuations (Right). 
When RKL-based supervision is combined with the RLVR framework, as in KDRL, the instability in response length and gradient norm is partially mitigated. 
However, its training reward remains lower than the pure RLVR baseline GRPO++, suggesting that RKL-based OPD still negatively affects RLVR optimization. 
In contrast, TGPO converges stably with controlled response lengths while achieving stronger final benchmark performance than GRPO++, as shown in Table~\ref{tab:main_results}. 
These results suggest that TGPO effectively combines on-policy exploration with teacher supervision under large teacher--student divergence.

Finally, although LUFFY appears to achieve the highest training reward, this value is likely inflated by its strategy of including a ground-truth sample in each training group, which may also lead to response length instability and large gradient norm fluctuations observed in Figure~\ref{fig:train_dynamics}.

\textcolor{lxy}{\subsection{Top-1 Guidance Ablation}
\label{app:fkl_ablation}
To examine whether retaining more of the teacher's probability mass improves the guidance signal, we compare TGPO's top-1 target with larger top-$k$ approximations and a Monte-Carlo-sampled teacher target, using \texttt{Qwen2.5-Math-1.5B} as the student and \texttt{Qwen3-30B-A3B} as the teacher (the Cross-Family setting of Section~\ref{sec:rkl-limitations}). 
All variants share the same student, training data, optimization setup, and teacher forward passes, differing only in the per-token guidance target. We evaluate every 20 training steps and report the best average test score across evaluation steps, with each score averaged over AIME, AMC, MATH, Minerva, and OlympiadBench.}

\begin{table}[t]
\centering
\small
\setlength{\tabcolsep}{6pt}
\begin{tabular*}{\columnwidth}{@{\extracolsep{\fill}}lc@{}}
\toprule
\rowcolor{white}
Teacher target & Best avg. test score \\
\midrule
Top-1 (TGPO) & \textbf{0.336} \\
Top-5 & 0.330 \\
Top-10 & 0.348 \\
Top-100 & 0.341 \\
Monte-Carlo & 0.345 \\
\bottomrule
\end{tabular*}
\caption{Comparison of forward-KL teacher targets under a matched 100-step training budget. Scores are the best average test accuracy across AIME, AMC, MATH, Minerva, and OlympiadBench.}
\label{tab:fkl_ablation}
\end{table}

\textcolor{lxy}{Table~\ref{tab:fkl_ablation} shows no consistent improvement from retaining more teacher probability mass. 
All variants achieve similar performance, with scores ranging from 0.330 to 0.348, while the top-1 target used by TGPO achieves 0.336. 
Together with the top-$k$ coverage analysis in Appendix~\ref{app:topk_coverage}, these results show that the additional teacher probability mass does not translate into better performance. 
We therefore adopt the top-1 target in TGPO, which provides comparable performance at the lowest supervision cost.}

\subsection{TGPO with Different Teachers}
\label{sec:diff_teachers}

\begin{table}[t!]
    \centering
    \resizebox{\linewidth}{!}{%
        \begin{tabular}{lccccc}
            \toprule
            \rowcolor{white}
            \textbf{Teacher Model} & \textbf{AMC} & \textbf{MATH} & \textbf{Olympiad}  & \textbf{GPQA}$^{*}$ & \textbf{Avg.} \\
            \midrule
            No Teacher           & 58.3 & 82.2 & 47.3 & 32.3 & 55.0 \\
            R1-Distill-Qwen-32B  & 57.8 & 83.4 & 47.4 & \textbf{40.9} & 57.4 \\
            Qwen3-30B-A3B        & \textbf{60.2} & \textbf{84.4} & \textbf{49.8} & 37.4 & \textbf{58.0} \\
            \bottomrule
        \end{tabular}%
    }
    \caption{Ablation study on different teacher models. We compare the performance of TGPO when guided by different teacher policies with a no-teacher baseline.}
    \label{tab:ablation_teacher}
\end{table}

To evaluate whether TGPO generalizes across different teacher models, we compare a baseline trained without teacher guidance (No Teacher) against TGPO variants guided by \model{R1-Distill-Qwen-32B} and \model{Qwen3-30B-A3B}. 
As shown in Table~\ref{tab:ablation_teacher}, incorporating teacher guidance consistently improves performance over the pure RLVR baseline. 
TGPO with \model{Qwen3-30B-A3B} achieves the best average accuracy (58.0\%) and performs particularly well on mathematical benchmarks, including AMC, MATH, and Olympiad. 
In contrast, TGPO with \model{R1-Distill-Qwen-32B} obtains the strongest result on GPQA (40.9\%). 
These results suggest that TGPO can effectively transfer the strengths of different teacher models and does not rely on a specific teacher architecture. We leave the exploration of a broader range of teacher models to future work.

\subsection{Comparison in the In-family Setting}

\begin{figure}[t!]
    \centering
    \includegraphics[width=0.92\linewidth]{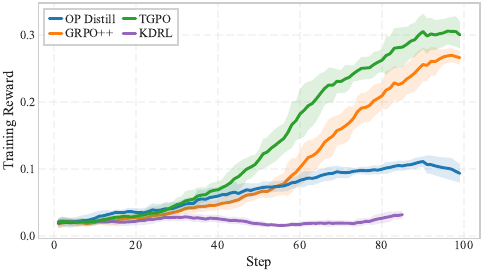}
    \caption{Training reward curves in the in-family setting. TGPO consistently achieves higher rewards than OP Distill, KDRL and GRPO++ throughout training.}
    \label{fig:infamily_compare}
\end{figure}

We further evaluate whether TGPO maintains its advantages in the in-family setting.
Following the setup in Section~\ref{sec:rkl-limitations}, we use \texttt{Qwen2.5-Math-7B} to supervise \texttt{Qwen2.5-Math-1.5B} for 100 training steps. 
We compare the on-policy distillation methods OP Distill, KDRL, and TGPO with the pure RLVR baseline GRPO++ by tracking the training reward.
As shown in Figure~\ref{fig:infamily_compare}, OP Distill achieves stable reward improvements, indicating that RKL-based OPD remains effective when the teacher and student distributions are relatively aligned.
GRPO++ obtains higher rewards, likely because outcome-based rewards better align with mathematical reasoning tasks. 
Although both OP Distill and GRPO++ improve steadily on their own, KDRL fails to converge in the in-family setting. 
In contrast, TGPO exhibits the fastest reward growth and consistently outperforms the other methods. 
These results show that TGPO remains effective beyond large-divergence regimes and generalizes well to in-family distillation scenarios.

\subsection{Impact of Guidance Scheduling}
\label{subsec:weight_effects}

\begin{figure}[t!]
    \centering
    \includegraphics[width=0.92\linewidth]{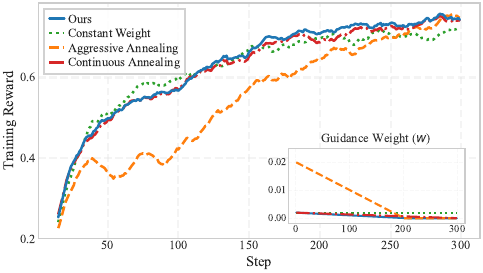}
    \caption{Ablation of annealing schedules. The inset details the guidance weight ($w$) schedule for each setting. Our method yields the best convergence.}
    \label{fig:ablation_schedule}
\end{figure}

To evaluate the guidance weight decay schedule introduced in Section~\ref{sec:integration}, we compare our strategy with three alternatives that use different initial guidance weights $w_{\text{init}}$ and decay rates $\delta$:
(1) Constant Weight ($w_{\text{init}}\!=\!2e\!-\!3$, $\delta\!=\!0$); (2) Aggressive Annealing ($w_{\text{init}}\!=\!2e\!-\!2$, $\delta\!=\!1e\!-\!4$); (3) Continuous Annealing ($w_{\text{init}}\!=\!2e\!-\!3$, $\delta\!\approx\!6.7e\!-\!6$, decaying to zero at the final training step); and (4) Ours ($w_{\text{init}}\!=\!2e\!-\!3$, $\delta\!=\!1e\!-\!5$, decaying to zero at step 200).

Figure~\ref{fig:ablation_schedule} shows that our schedule achieves the best overall performance. 
Aggressive Annealing suppresses rewards early in training, indicating that overly strong guidance limits exploration. 
Constant Weight performs competitively at first but plateaus early, suggesting that persistent imitation constraints hinder further reward optimization. 
Our method also outperforms Continuous Annealing, indicating that entering a pure RL phase before the end of training is important for effective policy optimization. By removing teacher guidance at step 200, our method achieves the highest final reward.

\section{Related Work}

\textcolor{lxy}{\paragraph{On-Policy Distillation.}
MiniLLM~\cite{gu2023minillm} introduced on-policy distillation (OPD) by sampling directly from the student distribution and applying reverse-KL (RKL) supervision (Eq.~\ref{eq:rkl}).
GKD~\cite{agarwal2024policy} subsequently unified forward- and reverse-KL distillation within a general on-policy framework and studied their combination with reinforcement learning objectives.
In particular, GKD favors reverse KL or JSD with a high reverse-KL weight for RL-oriented training, motivated by its compatibility with standard RL workflows.
KDRL~\cite{xu2025kdrl} explored integrating RKL-based distillation into RLVR through reward shaping and differentiable teacher regularization.
The \textit{On-Policy Distillation} blog by Thinking Machines~\cite{lu2025onpolicydistillation} further compared the training cost of OPD and SFT-then-RL pipelines, highlighting the potential of OPD as a post-training approach.
More recently, OPD has been extended to self-distillation settings such as OPSD~\cite{zhao2026self} and SDPO~\cite{hubotter2026reinforcement}, which use previous trajectories with error feedback to guide exploration.}

\textcolor{lxy}{Our work is closely related to GKD~\cite{agarwal2024policy}, as TGPO's guidance objective can be viewed as a hard-label (top-1) special case of on-policy forward-KL distillation. GKD primarily studies OPD under relatively small teacher--student distribution gaps and adopts a reverse-KL formulation when integrating distillation with RL. We instead investigate the large-divergence regime, where RKL-based supervision can become ineffective, and show that forward-KL provides effective directional teacher guidance.}


\paragraph{Discussion over Mixed-Policy.} In the context of LLM distillation, mixed-policy approaches~\cite{yan2025learning, zhang2025onpolicyrlmeetsoffpolicy}, which leverage samples from the teacher distribution, have achieved competitive results. However, our work remains strictly focused on the on-policy setting. We posit that on-policy learning, by optimizing the student's generation trajectory, offers greater robustness against distribution mismatch and ensures theoretical consistency with standard RL algorithms. By adhering to a strict on-policy setting, our insights are designed to not only advance LLM distillation but also generalize to fundamental RL research.

\section{Conclusion}
We present TGPO, an on-policy distillation method designed to overcome Reverse KL limitations. Compared to the sparse and uninformative signals provided by Reverse KL based algorithms, TGPO incorporates dense and explicit teacher guidance based on the student's rollout, while maintaining the robustness of on-policy learning. Empirical results on mathematical reasoning benchmarks demonstrate that TGPO not only outperforms baselines but also exhibits adaptability to various teacher models. Moreover, we demonstrate that applying guidance via differentiable regularization, coupled with a linear decay schedule, is essential for stable convergence and continued self-improvement. We hope our findings provide a theoretically grounded and practically effective direction for future advancements in LLM alignment.

\section*{Limitation}

Although TGPO demonstrates strong performance and improved training stability under large teacher--student policy divergence, the current framework is designed around the combination of token-level teacher guidance and trajectory-level verifiable rewards. As a result, TGPO is primarily suited to RLVR-style settings where reliable automatic verification signals are available. Its applicability to open-ended or subjective generation tasks remains less clear, particularly in scenarios where high-quality outcome reward models or rule-based verifiers are unavailable.
In addition, like most distillation-based methods, TGPO currently assumes access to a capable teacher model that can provide informative token-level supervision during training.
A promising direction for future work is to extend TGPO beyond verifiable reasoning tasks by incorporating stronger learned reward models or LLM-based judges, enabling more reliable supervision in domains with ambiguous or subjective evaluation criteria.

\section*{Acknowledgements}
This work was supported in part by the National Science Foundation of China (Nos. 62276056 and U24A20334), the Yunnan Fundamental Research Projects (No.202401BC070021), the Yunnan Science and Technology Major Project (No. 202502AD080014), the Liaoning Provincial Science and Technology Plan Project (No. 2026JH40/10100033), the Fundamental Research Funds for the Central Universities (Nos. N25BSS054 and N25BSS094), and the Program of Introducing Talents of Discipline to Universities, Plan 111 (No.B16009).

\bibliography{custom}

\newpage
\appendix

\section{Theoretical Analysis of RKL Instability}

\label{app:derivations}

In this appendix, we provide the formal derivations referenced in Section~\ref{subsec:low_reward}. We analyze the gradient behavior of the Reverse KL (RKL) objective and show why optimization becomes unstable when the student policy $\pi_\theta$ assigns probability mass to regions where the teacher policy $\pi_T$ has low probability, i.e., in the \textbf{Rejection} regime defined in Section~\ref{subsec:low_reward}.

\subsection{Gradient of RKL Objective}

Recall the RKL objective in Eq.~\ref{eq:rkl}:

\begin{equation*}
\begin{split}
    \mathcal{J}_{\text{RKL}}(\theta) &= \mathbb{E}_{x \sim \mathcal{D}} D_{\text{KL}}(\pi_\theta || \pi_T) \\
    &= \mathbb{E}_{x \sim \mathcal{D}, y \sim \pi_\theta(\cdot|x)} \left[ \log \frac{\pi_\theta(y|x)}{\pi_T(y|x)} \right].
\end{split}
\end{equation*}

For a fixed prompt $x$, let $J(\theta)\!=\!\mathbb{E}_{y \sim \pi_\theta} [\log \rho(y)]$, where $\rho(y)\!=\!\frac{\pi_\theta(y|x)}{\pi_T(y|x)}$ denotes the density ratio. Using the log-derivative trick, the gradient with respect to $\theta$ is:

\begin{align*}
\nabla_\theta J(\theta)
&= \nabla_\theta 
\mathbb{E}_{y \sim \pi_\theta}
\left[\log \rho(y)\right] \nonumber \\
&= \mathbb{E}_{y \sim \pi_\theta}
\Big[
\nabla_\theta \log \pi_\theta(y|x)
\cdot \log \rho(y)
\nonumber \\
&\hspace{1.5cm}
+ \nabla_\theta \log \rho(y)
\Big] \nonumber \\
&= \mathbb{E}_{y \sim \pi_\theta}
\Big[
\nabla_\theta \log \pi_\theta(y|x)
\cdot \log \rho(y)
\nonumber \\
&\hspace{1.5cm}
+ \nabla_\theta \log \pi_\theta(y|x)
\Big] \nonumber \\
&= \mathbb{E}_{y \sim \pi_\theta}
\left[
\nabla_\theta \log \pi_\theta(y|x)
\cdot (\log \rho(y)+1)
\right].
\label{eq:rkl_gradient}
\end{align*}

Note that the term resulting from $\mathbb{E}[\nabla_\theta \log \pi_\theta]\!=\!0$ is often omitted, but strictly speaking, the gradient is weighted by the term $(\log \rho(y) + 1)$.


In the context of RL with intrinsic rewards (as discussed in Section~\ref{subsec:low_reward}), the RKL term serves as a negative reward, $- \log \rho(y)$. Under the policy gradient framework, the resulting stochastic gradient estimator $\hat{g}(y)$ for a sampled trajectory $y$ is proportional to:
\begin{equation*}
\hat{g}(y) \propto \nabla_\theta \log \pi_\theta(y|x) \cdot \big(-\log \rho(y)\big).
\end{equation*}

\subsection{Instability in the Rejection Regime}

We now analyze the gradient behavior in the \textbf{Rejection} regime. 
Although language models generate tokens autoregressively (i.e., $\pi(y|x)\!=\!\prod_{t=1}^L \pi(y_t | y_{<t}, x)$), our analysis operates at the complete trajectory level $y$. 
This perspective is crucial because the density ratio accumulates over the sequence length, amplifying the variance.

\paragraph{Unbounded Gradient Scaling.}

Consider a ``bad'' sample $y_{\text{bad}}$ in the \textbf{Rejection} regime, where
$\rho(y) \gg 1$, i.e., $\pi_\theta(y|x) \gg \pi_T(y|x)$. This corresponds to trajectories that receive non-negligible probability under the student policy but are assigned near-zero probability by the teacher. Formally, assume $\pi_\theta(y_{\text{bad}}|x) \ge \delta$ for some constant $\delta > 0$, while $\pi_T(y_{\text{bad}}|x) \le \epsilon$ with $\epsilon \to 0$.

The log-density ratio is then lower bounded by:
\begin{equation*}
\begin{split}
\log \rho(y_{\text{bad}})
&=
\log \pi_\theta(y_{\text{bad}}|x)
-
\log \pi_T(y_{\text{bad}}|x)
\\
&\ge
\log \delta - \log \epsilon
=
\log \left(\frac{\delta}{\epsilon}\right).
\end{split}
\end{equation*}

As $\epsilon \to 0$, the term $\log(\delta/\epsilon)$ diverges to infinity. Consequently, the gradient scaling factor $|\log \rho(y_{\text{bad}})|$ can become arbitrarily large, inducing extremely high-variance gradient estimates and unstable optimization.

This issue is particularly severe in cross-family distillation, where architectural and reasoning-style discrepancies often cause the teacher to assign near-zero probability to otherwise plausible student trajectories.
This analysis directly explains the sharp gradient spikes observed in Figure~\ref{fig:sec2_compare} (Middle).

\paragraph{Variance Explosion.}

Optimization stability is closely related to the variance of the stochastic gradient estimator. 
A standard proxy for this variance is the second moment of the gradient norm:
\begin{equation*}
\mathbb{E}_{y \sim \pi_\theta}
\left[
\|
\nabla_\theta \log \pi_\theta(y|x)
\|^2
\cdot
(\log \rho(y))^2
\right].
\end{equation*}

In the \textbf{Rejection} regime, where $\pi_T(y|x) \to 0$, the log-density ratio $\log \rho(y)$ can become arbitrarily large. Consequently, the weighting term $(\log \rho(y))^2$ grows without bound, substantially increasing the second moment of the gradient estimator and inducing extremely high gradient variance.

Such variance can severely destabilize optimization, particularly for adaptive optimizers such as Adam, resulting in gradient spikes, noisy parameter updates, and degradation of previously learned capabilities.

\subsection{Asymmetry of Reward Scaling}

In Section~\ref{subsec:low_reward}, we argued that RKL induces an asymmetric optimization landscape.
We formalize this observation by analyzing the intrinsic reward
\[
r_{\text{int}}(y)
=
-\log \rho(y)
=
\log
\frac{\pi_T(y|x)}
{\pi_\theta(y|x)}.
\]

\begin{itemize}[topsep=0pt, partopsep=0pt, itemsep=2pt, parsep=0pt, leftmargin=*]

\item
\textbf{Positive Rewards are Probabilistically Suppressed (Consensus Regime):}
Large positive rewards arise when
$\pi_T(y|x) \gg \pi_\theta(y|x)$,
meaning the teacher assigns substantially higher probability to a trajectory than the student.
However, trajectories are sampled from the student policy $\pi_\theta$.
Thus, large positive rewards are associated with trajectories that are already unlikely to be sampled.
As $\pi_\theta(y|x) \to 0$,
the probability of observing such trajectories vanishes, making strong positive reinforcement events exceedingly rare in practice.

\item
\textbf{Negative Penalties are Frequent and Unbounded (Rejection Regime):}
Conversely, large negative rewards arise when
$\pi_\theta(y|x) \gg \pi_T(y|x)$.
Since trajectories are sampled from the student policy, such rejection trajectories are likely to be observed during optimization.
At the same time, the intrinsic reward
$r_{\text{int}}(y)$
becomes unbounded below as
$\pi_T(y|x)\to 0$.
Consequently, optimization is repeatedly dominated by large-magnitude negative updates.
This asymmetry---where positive rewards are rarely observed while negative penalties are both frequent and unbounded---drives the variance explosion and instability discussed above.

\end{itemize}

\section{Teacher--Student Divergence Analysis}
\label{app:divergence}

To quantify the degree of teacher--student policy divergence in our experiments, we evaluate both teachers on rollouts sampled from the same \model{Qwen2.5-Math-1.5B} student under teacher forcing. We consider two teacher configurations: (i) an \textit{In-Family} teacher, \model{Qwen2.5-Math-7B}, which belongs to the same model family as the student; and (ii) a \textit{Cross-Family} teacher, \model{Qwen3-30B-A3B}, which differs in both model family and reasoning regime. We quantify divergence using complementary token-level metrics, including teacher perplexity, top-$k$ agreement, and trajectory-level divergence. Unless otherwise specified, all statistics are computed over tokens in the student's sampled trajectories.

\paragraph{Token-level divergence.}
Table~\ref{tab:divergence_full} reports the detailed statistics. The Cross-Family teacher consistently exhibits greater divergence from the student across all metrics. Its perplexity on student rollouts is 22\% higher than that of the In-Family teacher (3.71 vs.\ 3.03), while reverse and forward KL divergences are approximately $1.6$--$2.2\times$ larger. The Cross-Family teacher also has lower top-$k$ overlap with the student, with the top-5 overlap decreasing from 0.719 to 0.635.

We additionally examine the density ratio
\begin{equation}
\rho_t =
\frac{\pi_{\mathrm{teacher}}(y_t \mid x, y_{<t})}
{\pi_{\mathrm{student}}(y_t \mid x, y_{<t})},
\end{equation}
where $y_t$ denotes the token sampled from the student policy. We report the dispersion of $\log\rho_t$ as a measure of how differently the two policies score student-generated tokens. The standard deviation increases from 0.772 for the In-Family teacher to 1.458 for the Cross-Family teacher, indicating substantially larger token-level disagreement.

\begin{table}[t]
\centering
\small
\setlength{\tabcolsep}{4pt}
\begin{tabular*}{\columnwidth}{@{\extracolsep{\fill}}lcc@{}}
\toprule
 & In-Family & Cross-Family \\
\midrule
Teacher PPL $\downarrow$ & 3.03 & 3.71 \\
Reverse KL (top-1) $\downarrow$ & 0.182 & 0.405 \\
Reverse KL (top-$k$) $\downarrow$ & 0.444 & 0.743 \\
Forward KL (top-$k$) $\downarrow$ & 0.468 & 0.762 \\
Token mismatch ratio $\downarrow$ & 0.218 & 0.246 \\
Top-1 overlap $\uparrow$ & 0.889 & 0.839 \\
Top-5 overlap $\uparrow$ & 0.719 & 0.635 \\
Top-10 overlap $\uparrow$ & 0.702 & 0.624 \\
\bottomrule
\end{tabular*}
\caption{Detailed teacher--student divergence metrics on student rollouts under teacher forcing. In-Family and Cross-Family denote the \model{Qwen2.5-Math-7B} and \model{Qwen3-30B-A3B} teachers, respectively.}
\label{tab:divergence_full}
\end{table}

\paragraph{Generation-length mismatch.}
Beyond token-level divergence, we examine output lengths under independent decoding as a coarse behavioral indicator of reasoning-regime mismatch. Table~\ref{tab:length_divergence} shows that the In-Family teacher produces outputs with a length distribution similar to the student, with mean lengths of 1083 and 1115 tokens, respectively. In contrast, the Cross-Family teacher produces substantially longer traces, with a mean length of 3946 tokens and a median of 3642 tokens. Its truncation rate at the 8192-token limit is 0.056, compared with 0.001 for the student, corresponding to a $56\times$ increase.

\begin{table}[t]
\centering
\small
\setlength{\tabcolsep}{5pt}
\begin{tabular*}{\columnwidth}{@{\extracolsep{\fill}}lrrrr@{}}
\toprule
Model & Mean & Median & P95 & Trunc. \\
\midrule
Student (1.5B) & 1115 & 664 & 4160 & 0.001 \\
In-Family (7B) & 1083 & 732 & 3742 & 0.002 \\
Cross-Family (30B) & 3946 & 3642 & 8192 & 0.056 \\
\bottomrule
\end{tabular*}
\caption{Output-length statistics under independent decoding. Truncation rates are measured at the 8192-token limit.}
\label{tab:length_divergence}
\end{table}

Overall, these measurements establish a substantially larger teacher--student policy divergence in the Cross-Family setting than in the In-Family setting. This provides a controlled basis for studying the behavior of RKL-based OPD under different levels of teacher--student divergence.

\section{Cross-Family OPD Across RKL Implementations}
\label{app:rkl_variants}

To examine whether the instability observed in the Cross-Family setting is specific to a particular implementation of RKL-based OPD, we evaluate three RKL formulations under the large teacher--student policy divergence: (i) the standard advantage form, which uses the sampled student token as a Monte-Carlo estimator; (ii) the regularization form; and (iii) a top-100 distributional approximation over the student's token distribution. We use \model{Qwen2.5-Math-1.5B} student and evaluate all variants within a matched $\leq 50$-step window.

\begin{table}[t]
\centering
\small
\setlength{\tabcolsep}{3pt}
\begin{tabular*}{\columnwidth}{@{\extracolsep{\fill}}lccc@{}}
\toprule
RKL variant & Max. len. & Blow-up & Test score \\
\midrule
Advantage (MC) & 8189 & $\sim$38 & 0.142 \\
Regularization ($k{=}1$) & 8189 & $\sim$39 & 0.177 \\
Top-100 & 8101 & $\sim$55 & 0.092 \\
\bottomrule
\end{tabular*}
\caption{RKL-based OPD under large teacher--student policy divergence. All three RKL variants exhibit length blow-up within 50--55 training steps. Test scores are the best average scores within the matched $\leq 50$-step window.}
\label{tab:rkl_variants}
\end{table}

\paragraph{Results.}
Table~\ref{tab:rkl_variants} summarizes the results. All three RKL variants exhibit the same qualitative failure: the mean response length rapidly grows from $\sim$1.2K tokens and reaches the 8192-token truncation limit within 38--55 training steps. We define \emph{length blow-up} as the first step at which the mean response length exceeds 8000 tokens. The advantage and regularization formulations use the direct $k1$ KL estimator and yield equivalent gradients, while the top-100 variant provides a substantially lower-variance distributional approximation. Nevertheless, all three variants exhibit rapid length growth and eventual truncation.

For comparison, forward-KL variants and TGPO maintain stable mean response lengths of $\sim1.2-1.3$K tokens throughout training. Within the matched $\leq 50$-step window, TGPO achieves a best average test score of 0.187, compared with 0.142, 0.177, and 0.092 for the advantage, regularization, and top-100 RKL variants, respectively. These results show that the observed instability persists across different RKL implementations, suggesting that it is not an artifact of the Monte Carlo estimator but is associated with the RKL direction under large teacher--student policy divergence.

\section{Top-$k$ Coverage Analysis}
\label{app:topk_coverage}
Since the full teacher distribution is expensive to evaluate over a vocabulary of $\sim$151K tokens, we approximate it using the teacher's top-$k$ support. To assess the fidelity of this approximation, we measure the probability mass captured by the top-$k$ support on student-visited states. We collect 36,585 such positions and compute the coverage as
\begin{equation}
\label{eq:topk_coverage}
C_k(s_t)
=
\sum_{v\in\mathcal{V}_k(s_t)}
\pi_T(v\mid s_t).
\end{equation}

Table~\ref{tab:topk_coverage} reports the coverage statistics. Top-1 captures 93.7\% of the teacher probability mass on average, while top-5 and top-10 capture 99.87\% and 99.98\%, respectively. Top-100 captures essentially all probability mass. Thus, top-$k$ supports provide a close approximation to the full teacher distribution on student-visited states.

\begin{table}[t]
\centering
\small
\setlength{\tabcolsep}{4pt}
\begin{tabular*}{\columnwidth}{@{\extracolsep{\fill}}lccc@{}}
\toprule
Support & Mean & Median & 5th Percentile \\
\midrule
Top-1   & 0.937  & $\sim$1.000 & 0.567 \\
Top-5   & 0.9987 & $\sim$1.000 & 0.9966 \\
Top-10  & 0.9998 & $\sim$1.000 & 0.9997 \\
Top-100 & $\approx$1.000 & $\sim$1.000 & $\approx$1.000 \\
\bottomrule
\end{tabular*}
\caption{Probability mass captured by the teacher's top-$k$ support on 36,585 student-visited positions.}
\label{tab:topk_coverage}
\end{table}

\section{Experimental Details}
\label{app:implementation}

In this appendix, we provide detailed experimental settings, hyperparameter configurations, and additional empirical results.

\subsection{Detailed Setup}
\label{app:exp_detail}

\paragraph{Training Dataset.}
All experiments use a unified dataset derived from a subset of OpenR1-Math-46k-8192. 
We retain the original prompts from NuminaMath 1.5 but reconstruct the reasoning traces to enable a controlled comparison across different training paradigms. 
Specifically, TGPO is an on-policy method that requires only prompts and generates rollouts during training. 
In contrast, the off-policy and mixed-policy baselines require static reasoning traces. To support these baselines under the same prompt distribution, we construct a shared set of teacher-generated trajectories using Qwen3-30B-A3B. We then validate the generated traces with Math-verify and retain only valid samples. The final curated dataset contains 34,975 prompts paired with verified teacher-generated reasoning traces for off-policy and mixed-policy training.

\paragraph{Training Configuration.}
In addition to Qwen2.5-Math-7B, we also evaluate our method on the smaller Qwen2.5-Math-1.5B model. For the main experiments, we use Qwen3-30B-A3B as the teacher model. This setting introduces a large capability gap between the teacher and the student, corresponding to the \textbf{Rejection} regime analyzed in our paper. For the in-family experiments in Section~\ref{subsec:prelim_exp}, we replace the teacher model with Qwen2.5-Math-7B while keeping all other settings unchanged. 
To ensure a fair comparison, all RL-based methods use a fixed sampling budget of $K=8$ rollouts per prompt. We use a constant learning rate of $1\times10^{-6}$ and train all RL models for 300 steps. All experiments are conducted on a cluster of 8 NVIDIA A100 GPUs.
Our implementation is based on the verl framework\footnote{https://github.com/verl-project/verl} and uses vLLM\footnote{https://github.com/vllm-project/vllm} for efficient rollout generation.

\paragraph{Model Configuration.}
The native context window of Qwen2.5-Math-7B and Qwen2.5-Math-1.5B (4,096 tokens) is insufficient to accommodate the long reasoning traces in the off-policy data. To address this issue, we modify the model configuration by increasing the RoPE base frequency ($\theta$) from 10,000 to 40,000 and extending the context window to 16,384 tokens. In contrast, Qwen3-30B-A3B already supports a sufficiently large context window, so we keep its RoPE configuration unchanged. In addition, we resize the vocabulary dimensions of the student and teacher models to the same size to ensure tokenizer compatibility during training.

\paragraph{SFT Implementation.}
For all SFT baselines, we use the same dataset of prompts and Qwen3-30B-A3B-generated reasoning traces described above. We follow the training protocol of OpenR1~\cite{openr1}, which reproduces the performance of the distilled DeepSeek-R1 models. Specifically, we train each model for 3 epochs with a global batch size of 64 and a learning rate of $5\times10^{-5}$. We use a warmup ratio of 0.1 and set the maximum sequence length to 16,384 tokens.

\subsection{System Prompt}
\label{app:system_prompt}

\begin{tcolorbox}[
    colback=blue!5!white, 
    colframe=black,       
    arc=0mm,              
    boxrule=0.5pt,        
    left=10pt, right=10pt, top=10pt, bottom=10pt 
]
\small
Your task is to follow a systematic, thorough reasoning process before providing the final solution. This involves analyzing, summarizing, exploring, reassessing, and refining your thought process through multiple iterations. Structure your response into two sections: Thought and Solution. In the Thought section, present your reasoning using the format: ``\texttt{<think>}\textbackslash n thoughts \texttt{</think>}\textbackslash n''. Each thought should include detailed analysis, brainstorming, verification, and refinement of ideas. After ``\texttt{</think>}\textbackslash n'' in the Solution section, provide the final, logical, and accurate answer, clearly derived from the exploration in the Thought section. If applicable, include the answer in \texttt{\textbackslash boxed\{\}} for closed-form results like multiple choices or mathematical solutions.

\vspace{5pt}
\textbf{User:} \texttt{\{QUESTION\}} \\
\textbf{Assistant:} \texttt{<think>}
\end{tcolorbox}

\section{Results on the 1.5B Model}
\label{app:1d5b_exp}

\subsection{Overall Performance}
To further evaluate TGPO under a larger teacher--student capability gap, we conduct experiments using \model{Qwen2.5-Math-1.5B} as the student model and \model{Qwen3-30B-A3B} as the teacher model. We compare TGPO against GRPO++, RKL, KDRL, and LUFFY. Results are summarized in Table~\ref{tab:main_results_1.5b}.

TGPO achieves the best overall performance, reaching an average score of 33.5\% and outperforming both on-policy and off-policy baselines across most benchmarks. GRPO++ remains competitive with 32.2\% average accuracy but still falls short of TGPO, while LUFFY performs noticeably worse than its 7B counterpart. 

In contrast, RKL-based on-policy distillation methods become highly unstable in this setting. KDRL achieves only 4.7\% average accuracy, and both KDRL and RKL exhibit rapid training collapse, with generation lengths frequently saturating the maximum context window. We therefore report results from the best-performing checkpoints.

\begin{table*}[t]
\centering
\resizebox{0.75\textwidth}{!}{%
\begin{tabular}{lcccccc>{\columncolor{gray!20}}c} 
\toprule
\textbf{Model} & \textbf{AIME 24} & \textbf{AIME 25} & \textbf{AMC} & \textbf{MATH-500} & \textbf{Minerva} & \textbf{Olympiad} & \textbf{Avg.} \\
\midrule
Qwen2.5-Math-1.5B  
    & 7.2 & 3.6 & 26.4 & 28.0 & 9.6 & 21.2 & 16.0 \\
\midrule
LUFFY
    & 5.8 & 4.9 & 32.6 & 64.0 & 22.4 & 24.7 & 25.7 \\
GRPO++
    & 10.1 & 7.4 & 41.8 & 69.4 & 28.3 & 35.9 & 32.2 \\
KDRL
    & 0.9 & 0.3 & 5.9 & 11.4 & 5.1 & 4.7 & 4.7 \\
OP Distill
    & 1.7 & 0.4 & 18.8 & 45.2 & 16.2 & 14.7 & 16.2 \\
\midrule
\textbf{TGPO}
    & \textbf{11.8} & 7.8 & 43.0 & \textbf{71.4} & \textbf{30.5} & \textbf{36.6} & \textbf{33.5} \\
\bottomrule
\end{tabular}}
\caption{Performance evaluation based on Qwen2.5-Math-1.5B. The teacher model employed is Qwen3-30B-A3B. All models are evaluated under a unified setting. Bold indicates the best result.}
\label{tab:main_results_1.5b}
\end{table*}

\begin{figure*}[t!]
    \centering
    \includegraphics[width=\linewidth]{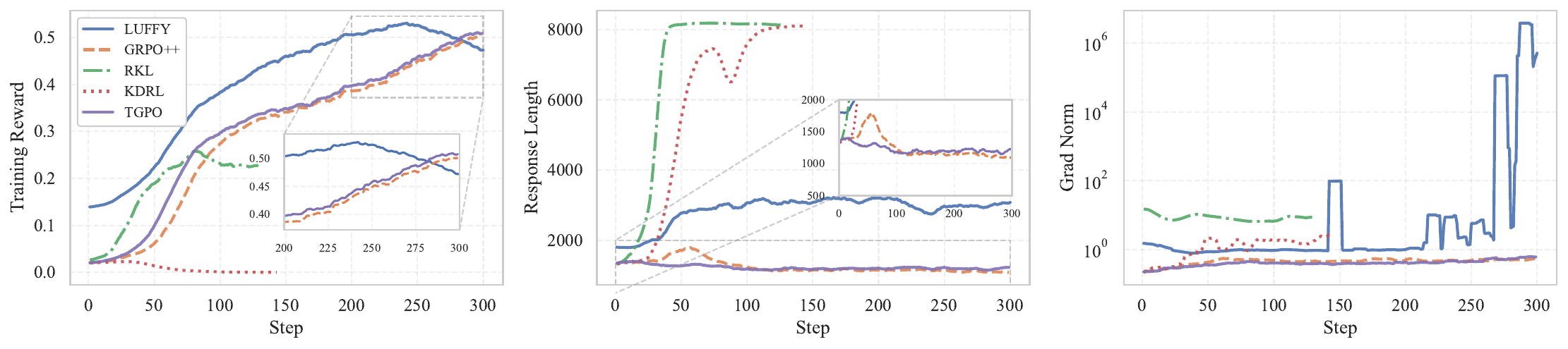}
    \caption{Training Dynamics Analysis. 
    \textbf{(Left)} Training reward. TGPO demonstrates robust growth and convergence compared to RKL and KDRL. 
    \textbf{(Middle)} Response length. TGPO avoids RKL's length explosion and aligns with GRPO++'s stability. 
    \textbf{(Right)} Gradient norm. TGPO shows stable optimization compared to the high variance in RKL, KDRL and LUFFY.}
    \label{fig:train_dynamics_1.5b}
\end{figure*}

\subsection{Training Dynamics}

To better understand the performance differences across methods, we analyze the training dynamics of the 1.5B student model. Figure~\ref{fig:train_dynamics_1.5b} shows the training reward, response length, and gradient norm during the first 300 optimization steps.

TGPO achieves stable reward improvement throughout training and converges to higher reward values than RKL-based methods, while KDRL shows almost no reward improvement from the beginning of training. In terms of response length, both RKL and KDRL quickly exhibit length explosion, with generation lengths saturating the 8,192-token rollout limit. TGPO avoids this behavior and maintains response lengths comparable to GRPO++. TGPO also maintains relatively stable gradient norms throughout training, whereas RKL and KDRL exhibit substantially higher variance and LUFFY shows several large gradient spikes. Overall, these results further support the instability of RKL-based supervision under large teacher--student capability gaps and show that TGPO maintains stable optimization behavior in this setting.


\end{document}